\documentclass[sigconf,screen]{acmart}

\renewcommand\footnotetextcopyrightpermission[1]{} % removes footnote with conference information in first column
\AtBeginDocument{%
  \providecommand\BibTeX{{%
    \normalfont B\kern-0.5em{\scshape i\kern-0.25em b}\kern-0.8em\TeX}}}

\setcopyright{acmcopyright}
\copyrightyear{2021}
\acmYear{2021}
\acmDOI{10.1145/1122445.1122456}

\usepackage{subfigure}
\usepackage{graphicx}
\usepackage{float}
\newtheorem{definition}{Definition}

\begin{document}

\title{A Data-Driven Column Generation Algorithm For Bin Packing Problem in Manufacturing Industry}

\author{Jiahui Duan}
% \authornote{note.}
\email{duanjiahui@stu.scu.edu.cn}
\affiliation{%
  \institution{Sichuan University}
  \institution{Huawei Noah's Ark Lab}
  \country{China}
}

\author{Xialiang Tong}
\email{tongxialiang@huawei.com}
\affiliation{%
  \institution{Huawei Noah's Ark Lab}
  \country{China}
}

\author{Fei Ni}
\email{nifei@hust.edu.cn}
\affiliation{%
  \institution{Huazhong University of Science and  Technology}
  \institution{Huawei Noah's Ark Lab}
  \country{China}
}
 \author{Zhenan He}
\email{zhenan@scu.edu.cn}
\affiliation{%
  \institution{Sichuan University}
  \country{China}
}
 \author{Lei Chen}
\email{lc.leichen@huawei.com}
\affiliation{%
  \institution{Huawei Noah's Ark Lab}
  \country{China}
}
 \author{Mingxuan Yuan}
\email{Yuan.Mingxuan@huawei.com}
\affiliation{%
  \institution{Huawei Noah's Ark Lab}
  \country{China}
}

\renewcommand{\shortauthors}{}

\begin{abstract}
The bin packing problem exists widely in real logistic scenarios (e.g., packing pipeline, express delivery), with its goal to improve the packing efficiency and reduce the transportation cost. In this NP-hard combinatorial optimization problem, the position and quantity of each item in the box are strictly restricted by complex constraints and special customer requirements. Existing approaches are hard to obtain the optimal solution since rigorous constraints cannot be handled within a reasonable computation load. In this paper, for handling this difficulty, the packing knowledge is extracted from historical data collected from the packing pipeline of Huawei. First, by fully exploiting the relationship between historical packing records and input orders(orders to be packed) , the problem is reformulated as a set cover problem. Then, two novel strategies, the constraint handling and process acceleration strategies are applied to the classic column generation approach to solve this set cover problem. The cost of solving pricing problem for generating new columns is high due to the complex constraints and customer requirements. The proposed constraints handling strategy exploits the historical packing records with the most negative value of the reduced cost. Those constraints have been implicitly satisfied in these historical packing records so that there is no need to conduct further evaluation on constraints, thus the computational load is saved. To further eliminate the iteration process of column generation algorithm and accelerate the optimization process, a Learning to Price approach called Modified Pointer Network is proposed, by which we can determine which historical packing records should be selected directly. Through experiments on real-world datasets, we show our proposed method can improve the packing success rate and decrease the computation time simultaneously.

\end{abstract}





\maketitle

\section{Introduction}

To meet the order requirement from all over the world, Huawei handles millions of orders every month.  The processing of each order involves several necessary procedures including production, packing, and transportation, among which the packing process is required to put all needed items into specific smallest Pack Transport Unit (SPUs). Making superior packing plan is of great significance in building logistic system since the slight improvement of packing quality can always bring  significant reduction of cost especially when the number of orders is huge.  In real application, the feasible solution of packing task can be obtained by solving Three-Dimension Bin Packing (3DBP) problem and its variants, in which items are restricted by some basic constraints, e.g., volume constraints, loading capacity constraints and overlap constraints.

However, in real packing scenario, besides above constraints, some special requirements from customers also need to be taken in consideration, e.g.,  customers specify the type of SPUs for their orders or request some items to be contained in the same SPU with a specific quantity ratio. In addition, incomplete information (e.g., the value of length, width, and height) of items and SPUs also aggravate the difficulty. In this case, the price of exact algorithms(e.g., Branch and Bound \cite{chen1995analytical}) is pretty high and the conflict between quality and solving time is hard to be balanced even if it spends huge computational resources. In recent years, heuristic approaches attract more attention from both academic and industry field. This type of method can generate a feasible solution within an acceptable time, however, without guaranteeing the optimality.

As for the practice of Huawei, in early stage,  it takes a lot of workers to generate packing plans manually, where the packing  success rate and the quality of packing result seriously depended on the experience of these workers. Our previous work \cite{chen2019data} constructed the packing framework named fuzzy matching based on the historical packing record. In this framework, items are packed stage by stage. At each stage, a local best SPU is chosen from the history record based on the current left items and then items contained in this SPU are packed.  At last, a packing plan can be obtained when there is no local best SPU can be chosen or all items have been packed. After fuzzy matching, a heuristic approach \cite{li2018data} is adopted for orders which have not been successfully packed.   The fuzzy matching approach helps output the packing plan within a short time and increase the packing success rate. However, there are some shortcomings in this approach. First, in this framework, the heuristic approach and history data are separated into two independent part. Second, the approach cannot fully excavate the potential of history apcking data for just finding the local best SPU in each stage, which may cause a low quality and even an infeasible plan.

In this paper, we propose a data driven column generation approach incorporating two novel strategies to address above issues. At the beginning, the huge number of history SPUs are collected from the real world, whose feasibility and quality are tested in practice. For each order, we find $\mathbf{Matched\ SPUs}$ from history packing records. With these $\mathbf{Matched\ SPUs}$, a set cover problem is formulated to exactly connect  historical SPUs and the order to be packed. By adding extra artificial variables, heuristic approach is incorporated to obtain an initial solution quickly. Then, the column generation algorithm \cite{desaulniers2006column} with proposed constraints handling and accelerating strategies is adopted to solve the problem. First, instead of calculating the pricing model directly, constraint handling strategy evaluates the reduced cost of historical SPUs and select the best one to be added to the model, by which the computational cost of solving pricing problem to generate a new column is avoided. To be noted, the difficulty of solving pricing problem are from mentioned constraints and customer requirements, however, these historical SPUs are with guaranteed feasibility under complex constraints and can be used directly as new columns for solving process.
%  Distinguished from the the traditional column generation which need pay huge computational cost in solving pricing problem which contains mentioned constraints to generate a new column at each iteration, we directly calculate the reduce cost of historical SPUs and select the best one into model. 
In addition, to further eliminate the iteration process of column generation approach and accelerate the solving process, a learning to price approach is proposed to predict the selected SPUs directly. This approach, named Modified Pointer Network (MPN), is constructed based on the Sequence to Sequence (Seq2Seq) architecture \cite{sutskever2014sequence}, in which a dynamic input scheme is embedded for handling the varied length of input and output.

The contributions of this paper are summarized as follows:
\begin{itemize}
\item With fully extracting the packing knowledge from historical packing records, we reformulate the 3DBP problem as a set cover problem which bridges the gap between the historical data and input orders. Meanwhile, the heuristic approach is integrated in this framework to find a feasible packing plan. 
\item A data driven column generation approach is adopted to solve the problem, in which a constraint handling strategy and a learning to price approach are developed. Since the constraints and customer requirements have been satisfied in historical SPUs, the constraints handling strategy select SPU from historical packing records based on the evaluated reduced cost, thus the computational cost of calculating exact pricing model is saved.  Meanwhile, our proposed MPN model with a dynamic input scheme attempts to learn to price the existed SPUs, thereby determining all SPUs to be used. 
\item Our proposed methods show outstanding performance compared to the existing algorithms. The experiment results based on real-world datasets, shows our approach could improve the packing success rate and computation time significantly. 
\end{itemize}
\section{Related work}
This part reviews the heuristic approaches and learn to optimization approaches. As the combinatorial NP-hard problem \cite{korte2012combinatorial}, according to the dimension, the Bin Packing Problem can be categorized into One-Dimension-Bin-Packing  \cite{wei2020new}, Two-Dimension-Bin-Packing \cite{blum2013solving}, and Three-Dimension-Bin-Packing \cite{korte2012combinatorial}. Meanwhile, some variants of Three-Dimension-Bin-Packing (3DBP) problem have been focused, e.g., 3DBP problem with variable height \cite{wu2010three}, 3DBP problem with item fragmentation \cite{menakerman2001bin}, etc.

\subsection{Heuristic Approach}
 Branch and Bound (B\&B) based framework \cite{chen1995analytical,martello2000three} has attempted to exactly solve the 3DBP problem. However, this type of solver can only be used in small scale problems with only a few items. Then, heuristic algorithms attract much attention for their ability to obtain a sub-optimal solution within acceptable solving time. Some of them focus on designing packing strategies including Extreme-Points based heuristic \cite{crainic2008extreme}, Best-fit approach \cite{johnson1974worst}, and the wall building algorithm \cite{george1980heuristic}. In addition, Evolutionary Computation (EC), a class of nature-inspired and population based approach, are also adopted to solve 3DBP problem, e.g., Genetic Algorithm \cite{gonccalves2013biased,kang2012hybrid}, Ant Colony Algorithm \cite{silveira2013aco}, and Particle Swarms Optimization \cite{li2017three}. Recently, another type of method  adopts Machine Learning approach \cite{hu2017solving, lopez2013understanding} to solve the 3DBP problem, these approaches mainly focus on training models to select heuristic strategy from some candidate heuristic or evaluating the quality of solutions.

Our previous work \cite{chen2019data} proposed a fuzzy matching approach which use historical data to pack items into SPUs while satisfying the constraints from classical 3DBP as well as customer demand. For each order, the history SPUs whose items are the subset of order's items are first selected as Matched SPUs. In each iteration, the best SPU is selected from all these matched SPUs based on real-life constraints and current left items. This process continues until all the items is packed or all SPUs cannot satisfy constraints.

In this paper, on basis of \cite{chen2019data}, we combine the heuristic approach and the historical data, reformulating the problem as a set cover problem to further improve the packing success rate.

\subsection{Learn to Optimize Approach}
The first type of learning approach is directly constructing optimal solutions. These approaches are always based on Seq2Seq network which was first proposed in \cite{sutskever2014sequence}. Then, the Pointer-Network \cite{vinyals2015pointer} changed the attention scheme on TSP problem. 
% Recently, Pointer-Network based approaches get success in solving routing problems. 
In \cite{DBLP:conf/iclr/BelloPL0B17}, the Reinforcement Learning is used for training Pointer Network on TSP problem. Research \cite{nazari2018reinforcement} proposed a modified attention mechanism focusing on solving VRP and its variants. In \cite{kool2018attention}, a self-attention approach was incorporated into Seq2Seq framework on several routing problems. Besides, Seq2Seq based approaches also show outperforming ability in Discrete Choice Modeling \cite{mottini2017deep} and Max-Cut Problem \cite{gu2020deep}.

The Seq2Seq architecture shows its superiority in optimization task for its ability to handle the sequence input, which always occurs in combinatorial optimization problem. Considering this reason, in this paper, we also adopt the similar network design.  

Another approach is embedded in traditional optimization framework e.g., Branch and Bound \cite{he2014learning} as well as Branch and Cut  \cite{tang2020reinforcement}.  The performance of these approach always depends on the choice of the variable to be branch, the node to be expanded, and the constraint to be cut. Research \cite{he2014learning} focused on selecting the node on branch tree and determining whether this node should be expanded or pruned, which decreased the solving time while guaranteeing the optimality. 
% \textcolor{red}{This approach helps Solver decide the sequence of node and decide quickly whether this node should be expanded, which decrease the solving time.}
Then, \cite{khalil2016learning,balcan2018learning} proposed a learning approach determining which variable should be branch in the current node.  Based on the Branch and Cut, \cite{tang2020reinforcement} proposed a Reinforcement Learning approach for generating Gomary Cut to tight the search space.

In this paper, we incorporate the learning approach in column generation framework and eliminate the computation cost from solving Pricing Problem. To our best knowledge, our proposed $\mathbf{Learning \ to \ Price}$ approach is the first attempt to combine  column generation and reinforcement learning.

\section{Problem Formulation}
In this section, we first illustrate the 3DBP task in Huawei supply chain. Then, through exploiting history data, we reformulate the 3DBP problem as a set cover problem.
\subsection{Three-Dimension-Bin-Packing Task}
Given an order $O_m=\left\{(t_i,q_i)|i=1,2,...D \right\}$ with extra customer demand  $C_m$, where $D$ denotes the number of item types,  $t_i$ is the $i$th item, and $q_i$ is its quantity, our task is packing all items into a set of SPUs. 
% For a SPU, $SPU_j$, the volume is $Q_j$ and the maximum load  capacity is $W_j$. 
During the packing process, for each SPU, the following hard constraints must be satisfied. First, the total weight of packed items  must be less than the load capacity of this SPU, and the total volume of packed items must be less than the volume of this SPU. Then, items inside the same SPU cannot overlap in any dimensions. In addition, some special constraints ($C_m$), from customer preference also need to be emphasized. For instance, the type of SPUs is specified by a part of orders, or some items are required to be in one SPU with a deterministic ratio. These special constraints cannot be formulated clearly in literature. 

Then, we present the history record as: 
\begin{align}
    record_{Aug}=\left\{SPU_1,SPU_2,...,SPU_R\right\}
\end{align} %权宜之计
where $record_{Aug}$ is the historical record containing $R$ packed SPUs from August. Each SPU in $record_{Aug}$ can be denoted as $SPU_p=\left\{(St_i,Sq_i)|i=1,2,...SD\right\}$, $p\in \left\{1,2,...R\right\}$,  which represents that it contains $SD$ types of items and the amount of $i$th item, $St_i$, is $Sq_i$. In addition, we denote the special related customer demand of $SPU_p$ to be $SC_p$. Naturally, we have definition as follows:
\begin{definition}[Matched SPU]
Given an order $O=\left\{(t_i,q_i)|i= 1\right.$ $\left.,2,...,D \right\}$ with the special demand $C$,  and a $SPU=\left\{(St_j,Sq_j)|j=1,\right.$ $\left.2,...,SD\right\}$ related to the demand $SC$, where $D$ and $SD$ are the number of item types in order and SPU, respectively. In this order, $t_i$ is the $i$th item and $q_i$ is its quantity. In $SPU$, the amount of the $j$th item ($St_j$) is $Sq_j$.

If $\forall j \in \left\{1,2,...,SD\right\}$, $\exists i \in\left\{1,2,...,D\right\}$, $t_i=St_j,\ q_i>=Sq_j$, and $SC=C$ are satisfied, then this SPU is a Matched SPU for given order $O$.

% there exists $SPU_p$ satisfying that any item $St_i$ in $SPU_p,p\in\left\{1,2,...R\right\}$ also exists in $O_m$ (e.g., $t_i$), and $Sq_i\leq q_i$, we define $SPU_p$ to be a matched SPU for $O_m$ if condition $SC_p=C_m$ can be satisfied. 
\end{definition}
For example, the order $O_m=\left\{(A,2),(B,4),(C,5)\right\}$ with a special demand $C_m$, which is the quantity ratio between item $B$ and item $A$ is two.  Suppose that there exist three historical SPUs which include $SPU_1=\left\{(A,1),(B,2)\right\}$, $SPU_2=\left\{(A,1),(D,2)\right\}$, and $SPU_3=\left\{(A,1),(B,3)\right\}$. Based on above definition, only $SPU_1$ can be selected as $\mathbf{Matched \ SPU}$.

\subsection{Problem Reformulation Based on Historical Data}
By fully taking utilization of the history data, we can reformulate the problem as a set cover problem. First, for the order $O_m=\left\{(t_i,q_i)|i=1,2,...D \right\}$ with $D$ types of items, we find all the matched SPUs, $S_m=\left\{ SPU_1,SPU_2,...SPU_P\right\}$, where the number of matched SPUs for $O_m$ is $P$. Then, the set cover problem can be stated as:
\begin{flalign}
\qquad & min \sum_{i=1}^{P}c_i\lambda_i \label{objective1} & \\
\qquad & \sum_{i=1}^{P}  Sq_{ji}\lambda_i = q_j, \ \forall j=1,2...,D \label{demand1} &\\
\qquad &  \lambda_i \ \in \mathbb{Z}^+, \ \forall i =1,2,...,P &
\end{flalign}
where $c_i$ and $Sq_{ji}$ are the cost of SPU and the amount of $j$th item contained in $SPU_i$, respectively. The integer variable $\lambda_i$ represents the  amount of $SPU_i$ used. The objective \eqref{objective1} minimizes the total cost of SPUs.
The set cover constraint \eqref{demand1} requires that the demand of each item in order must be satisfied exactly. To be noted, we adopt the set cover constraint \eqref{demand1} rather than the  set overcover constraint (e.g., $\sum_{i=1}^{N}  Sq_{ji}\lambda_i+\mu_j \geq q_j \ j=1,2...,D$) because  the quality of history SPU cannot be guaranteed if there are some items are removed. The set cover constraint is harder to be satisfied than set overcover constraint, so that we cannot find a feasible solution only based on historical data in some case. In addition, the number of $\mathbf{Matched \ SPUs}$ is always a huge value. Thus, solving this Interge Programming Model (IP) is also expensive. 

Therefore, to quickly get an initial solution, we incorporate the heuristic approach into this model. First, we randomly select a set of SPU (e.g, $N$ matched SPUs) from  $S_m$. Then,  the extra variable $\mu_j$ is artificially created, which is the amount of $j$th item fed into heuristic approach.   The problem is reformulated as :
\begin{flalign}
\qquad & min \sum_{j=1}^{D}\mu_j \label{objective2} & \\
\qquad & \sum_{i=1}^{N}  Sq_{ji}\lambda_i+\mu_j = q_j, \ \forall j=1,2...,D \label{demand2} & \\
\qquad & \lambda_i \ \in \mathbb{Z}^+, \ \forall i =1,2,...,P & \\
\qquad & \mu_j \ \in \mathbb{Z}^+, \ \forall j =1,2,...,D &
\end{flalign}
where the objective \eqref{objective2} is from the demand of real-scenario application that the number of items fed into heuristic approach should be as less as possible. With this formulation, we can easily get support from the potential of historical data and the ability of heuristic approach to avoid the huge computational cost of solving exact 3DBP problem. The packing plan from the above model are used as initial SPUs (e.g., $M$ SPUs) of column generation approach to further improve the solution's quality. By relaxing the integer constraint, the Restrict Master Problem (RMP) is formulated as:
\begin{flalign}
\qquad &   min \sum_{i=1}^M c_i\lambda_i + \sum_{t=0}^T c_t\eta_t  \label{objective3} & \\
\qquad &   \sum_{i=1}^M Sq_{ji}\lambda_i+\sum_{t=0}^T Sq_j^t\eta_t = q_j, \ \forall j=1,2,...D   \label{demand3} & 
\end{flalign}
and the dual formulation of RMP (DRMP) can be formulated as:
\begin{flalign}
\qquad &    max \sum_{j=1}^D q_j \pi_j^T  \label{dual_obj} & \\
\qquad &  \sum_{j=1}^D Sq_{ij} \pi_j^T \leq c_i, \ \forall i=1,2,...,M & \\
\qquad &  \sum_{j=1}^D Sq_j^t \pi_j^T \leq c_t, \ \forall t=1,2,...,T & 
\end{flalign}
where $Sq_j^t$ is the amount of the $j$th item in SPU which is generated in the $t$th iteration, $c_t$ is the  cost of this SPU. New integer variable $\eta_t$ means that the amount of  SPU generated in the $t$th iteration is used.  It is noted that the current iteration is $T$ and there is no new generated SPU when $T=0$. At each iteration, the RMP or DRMP is solved and the dual variable is $\Pi^T = \left\{ \pi_1^T,\pi_2^T,...,\pi_D^T, \right\}$. This dual variable is used for constructing pricing problem whose objective is minimizing the reduced cost of a new generated SPU and can be stated as:
\begin{flalign}
\qquad & min \ c_T-\sum_{j=0}^D Sq_j^T \pi_j^T &
\end{flalign}
where the variable $Sq_j^T$ is an integer variable and the amount of the $j$th item packed in the current SPU. After exactly solving the pricing problem at the current iteration, we can get a packed SPU, which is $SPU_T=\left\{(t_j,Sq_j^T)| j=1,2,...,D\right\}$.
If the  value of reduced cost $c_T-\sum_{j=0}^D Sq_j^T \pi_j^T$ is less than zero, this SPU can be added into RMP at the next iteration because it  certainly improves the objective value of RMP \cite{desaulniers2006column}.  Then, the cost of this SPU ($c_T$) and the amount of each item ($Sq_j^T$) contained in this SPU will be the coefficient of $\eta_T$ in objective \eqref{objective3} and constraints \eqref{demand3}, respectively. On the other hand, if the reduced cost of SPU obtained from pricing problem  is larger than zero, the iteration process stops. In this case, we add the integer constraints for RMP model and solve it to get a packing plan.

However, the pricing problem is actually a 3DBP problem  considering one SPU and a part of items, which means that if we directly solving the problem, we need handle those complex constraints. In addition, the huge number of iterations brings more computational load for us. Therefore, in the next section, we illustrate the approach proposed for addressing both two issues. 
\section{Proposed Method}
In this section, we first give a simple but effective constraints handling approach, which exploits the historical data and then helps us avoid the expensive computational cost of solving pricing problem. Then, to further accelerate the solving process, we propose a Modified Pointer Network (MPN) for learning to price the historical SPU. By selecting SPUs and add into RMP model, the iteration process of column generation is eliminated.
\subsection{Constraints Handling Strategy}
As mentioned above, at each iteration of column generation approach, we first solve the RMP and construct a new pricing problem based on dual variables. Then, solve this pricing problem and generate a new SPU which will be added in RMP. This process continues until we cannot obtain a packed SPU from pricing problem whose reduced cost is less than 0. Using historical data, at each iteration, we can directly computing the reduced cost of candidate SPUs which are not in RMP model:
\begin{flalign}
\qquad & \delta_i = c_i-\sum_j^D Sq_{ij}\pi_j^T \label{reducecost} &
\end{flalign}
where $c_i$ is the cost of the $i$th SPU and $Sq_{ij}$ is the amount of the $j$th items contained. $\Pi^T = \left\{ \pi_1^T,\pi_2^T,...,\pi_D^T, \right\}$ is the currently dual variable. Then, suppose that there are $N$ matched SPUs have not been added in RMP, we select the $i^*$th SPU if it satisfies:
\begin{flalign}
\qquad & i^* = \arg\min_{i} \delta_i, \ \forall i=1,2...,N &\\
\qquad & \delta_{i^*}<0 &
\end{flalign}

Thus, in each iteration, the computational complexity of selecting the SPU is $O(N)$, which is much smaller than exponential complexity  from exactly solving 3DBP probelm.

\subsection{Learning to Price}
\begin{figure*}[h]
  \centering
  \includegraphics[width=0.8\linewidth]{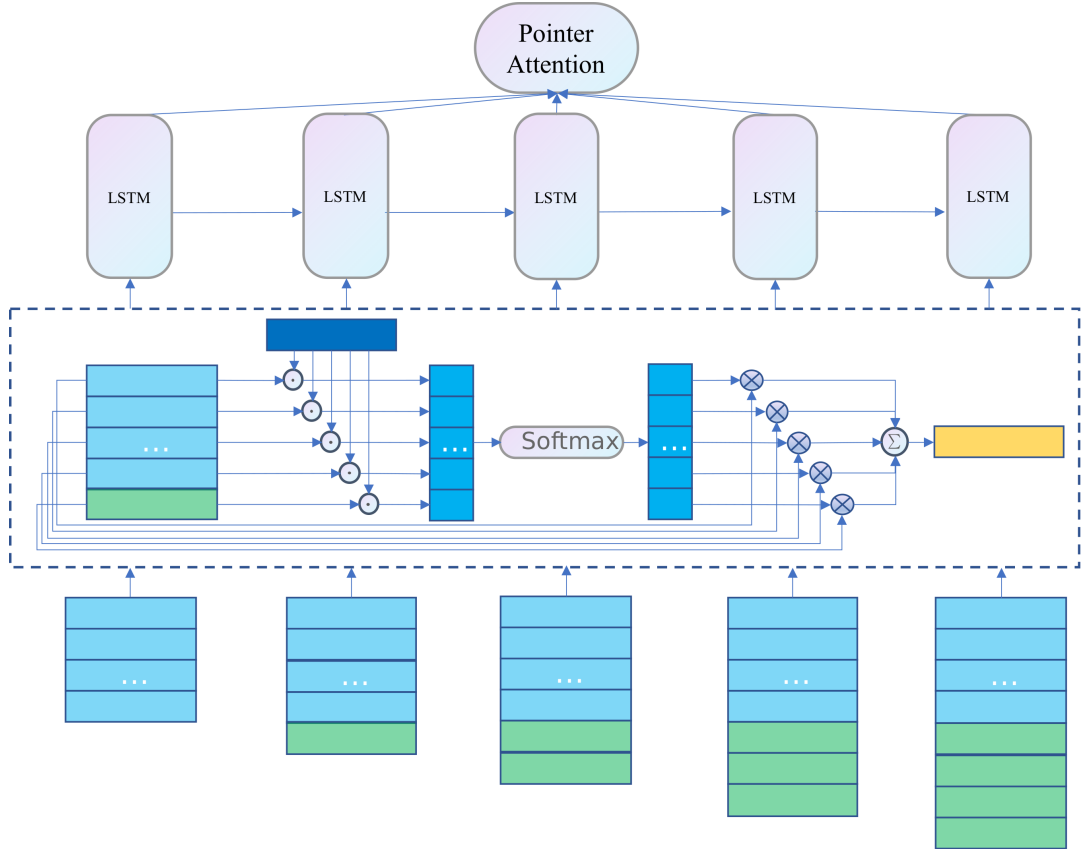}
  \caption{The Modified Pointer Network.}
  \Description{The picture with arrows}
  \label{MPN}
\end{figure*}
In this part, we illustrate  the detail of the proposed Modified Pointer Network (MPN), which selects SPUs from candidate to be added in RMP and thus eliminates the iteration process of column generation approach.

The first issue is that the number of items and the number of matched SPUs are various among orders, which means the length of input and output are varied. In this case, Pointer Network \cite{vinyals2015pointer} is an appropriate choice. The Pointer Network focuses on mapping a sequence input to a sequence output, but in our problem, the result is not affected by the sequence of SPUs. Therefore, the Read-Process-Write (Set2Set) \cite{VinyalsBK15}  architecture is a better choice for us. 

First, the embedding of the $i$th candidate SPU is denoted as $m_i$ and the embedding of the $i$th initial SPU in RMP is $im_i$, whose dimension is $1\times d$. The Process block takes the same action as \cite{VinyalsBK15}. The Write block is exactly a Pointer Decoder implemented by LSTM network \cite{hochreiter1997long} as shown in Fig.\ref{MPN}. At each step, Write block points to a SPU from candidates and this SPU is selected to be added into RMP. The  sampling probability distribution of pointing to a SPU $\tau_t$ is calculated:
\begin{flalign}
\qquad & u_{j}^t = V^T\tanh(W_1m_j+W_2d_{t}), \forall j =0,1,...,N \label{attention} & \\
\qquad &\tau_t \sim softmax(u^t+I^t)&
\end{flalign}
where $W_1$, $W_2$, and $V$ are trainable parameters. $d_t$ is the current hidden state from the Write block. The mask vector $I^t=\left\{I^t_{0},I^t_{1},...I^t_{N}\right\}$ prevents  Write block from pointing to the SPUs have been selected, where $I^t_{j}=-\infty$ if the $j$th SPU is selected before and $I^t_{j}=0$ otherwise. 

Distinguished from existed work that only fed the embedding of pointed SPU at each step, we  fed all SPUs already in RMP into Write block to simulate the status of RMP. However, the number of SPUs added in RMP varies during the SPU selecting process. To address this issue, we propose a dynamic input mechanism. Denote $SM^t=[im_1,im_2,...,im_M,m_1,...m_t]^T$ to be the embedding matrix of all SPUs already in RMP at the $t$th step, which include the initial SPUs and the pointed candidate SPUs. Therefore, the dimension of $SM^t$ is $(M+t)\times d$. The input of the Write block at the $t+1$th step can be calculated as:
\begin{flalign}
\qquad & a^t_i = f(SM^t_i, H), \forall i=1,2,...M,...M+t & \\
\qquad & e^t = softmax(a^t) & \\
\qquad & input^{t+1} = \sum_{i=1}^{M+t} e_i^tSM_i^t &
\end{flalign}
where $H$ is a trainable parameter with dimension $d\times1$. The function $f$ is a dot product function  calculating the weight of each embedding. By this approach, at each step, we can input all SPUs already added in RMP into Write block. Meanwhile, at each step, the $input^t$ is taken as $m_0$ in \eqref{attention}, which represents a stop signal of selecting process. Every time, when the Write block points to $m_0$, the MPN model stops and outputs all pointed SPUs. 

Next, we adopt the Reinforcement Learning as the training method:
\begin{itemize}
\item {\bf{Agent}}: the Write block in MPN is regarded as an agent.
\item {\bf{State}}: the state is directly defined as SPUs in RMP.
\item {\bf{Action}}: in each state, the action selects a SPU from candidates.
\item {\bf{Objective}}: the training objective is minimizing the negative expect reward:
\begin{flalign}
\mathcal{L}(\theta) = -\mathbb{E}_{\tau \sim p_\theta}[R(\tau)]
\end{flalign}
where $\theta$ is the parameter of MPN and $\tau$ is the set of selected  SPUs. The predicted SPUs are with swapping-invariance \cite{yang2019deep} which means the sequence of SPU is irrelevant to performance. Thus,   refereeing to \cite{yang2019deep}, the F1 score is adopted as reward function $R$ since it can measure the quality of unordered set.
\item {\bf{Self-critic}}: as stated in \cite{rennie2017self}, the self-critic approach adopts a reward obtained from an inference algorithm as baseline.  Therefore, the gradient can be formulated as :
\begin{flalign}
\nabla_\theta \mathcal{L}(\theta) = -[R(\tau)-R(\tau^g)]\nabla_\theta log(p_\theta(\tau))
\end{flalign}
where $\tau^g$ is the set of SPUs generated from an inference approach. In this paper, the greedy algorithm \cite{rennie2017self} is incorporated as this inference algotithm. Then, $p_\theta(\tau)$ is the probability of generating SPUs set $\tau$.
\end{itemize}

\section{Experimental Result}

\subsection{Data Set}
\begin{table*}
\caption{Features}
\label{features table}
\begin{tabular}{llc}
\toprule
Features   & Description  & Count \\ \midrule
Reduce cost                    & The value of the reduced cost calculated based on dual variable at the 0th iteration.                    & 1     \\
Stats. for RHS                & The right-hand-sides (RHS) for the item whose coefficient is   no-zero (min, max, mean).      & 3     \\
Stats. for dual                & The value of dual variables for the item whose   coefficient is no-zero (min, max, mean). & 3     \\
Stats. for coefficient          & The value of coefficient (min, max, mean, sum).                                     & 4     \\
Stats. for no-zero coefficient & The value of no-zero coefficient (count, min, mean).                                            & 3     \\
Stats. for To-demand            & The value of demand minus coefficient (min, max, mean, sum).                        & 4     \\
Stats. for Dual Times Coefficient         & The value of dual variables times   coefficient (min, max, mean).                  & 3     \\ \midrule
Total & & 21\\ \bottomrule
\end{tabular}
\end{table*}

\begin{table*}
  \caption{The Average Value of F1 Score}
  \label{F1 result}
  \begin{tabular}{ccccccccc}
    \toprule
    Order Source   & \multicolumn{2}{c}{8} & \multicolumn{2}{c}{9} & \multicolumn{2}{c}{10} & \multicolumn{2}{c}{11} \\
    History Record & 6         & 7         & 7         & 8         & 8          & 9         & 9          & 10        \\ \midrule
    MPN      & $\textbf{0.849}$	&$\textbf{0.862}$	&$\textbf{0.874}$	&$\textbf{0.968}$	&$\textbf{0.898}$	&$\textbf{0.876}$	&$\textbf{0.792}$	&$\textbf{0.881}$     \\
    MPN-W  & 0.806     & 0.704     & 0.744     & 0.924     & 0.637      & 0.764     & 0.592      & 0.736     \\
    RANDOM  & 0.487	&0.465	&0.444	&0.5	&0.482	&0.454	&0.416	&0.451     \\ \bottomrule
  \end{tabular}
\end{table*}

\begin{table*}
\caption{The Packing success Rate}
\label{rate}
\begin{tabular}{ccccc}
\toprule
Order Source    & 8     & 9 & 10 & 11 \\ \midrule
Fuzzy Match & 0.051 & 0.121 & 0.059  & 0.034  \\
Column Generation       & $\textbf{0.083}$ & $\textbf{0.217}$ & $\textbf{0.084}$  & $\textbf{0.048}$  \\ \bottomrule
\end{tabular}
\end{table*}

\begin{table*}
  \caption{Packing Quality and Solving Time }
  \label{packing result small}
 \begin{tabular}{ccccccccc}
 \toprule
Order Source             & \multicolumn{2}{c}{8} & \multicolumn{2}{c}{9} & \multicolumn{2}{c}{10} & \multicolumn{2}{c}{11} \\
Metric            & Ave.obj       & Total time      & Ave.obj       & Total time      & Ave.obj       & Total time      & Ave.obj        & Total time      \\ \midrule
Column Generation & $\textbf{8.72}$ & 71.35 & $\textbf{4.83}$   & 70.25& $\textbf{4.26}$& 60.21 & $\textbf{6.36}$ & 54.84                    \\
MPN                & 9.3    &$\textbf{44.86}$   & 5.01  & $\textbf{56.96}$ & 6.72                    & $\textbf{57.7}$                     & 6.53                    & $\textbf{47.19 }$                      \\ \bottomrule
\end{tabular}
\end{table*}
In this paper, all the experiment data are from the real-scenario of Huawei. For model training, input orders from August 2020 and historical SPUs from July 2020 are selected, in which $1\times10^4$ orders successfully packed with column generation approach are used for training, and $1\times10^3$ for evaluation. For each order, the number of items type and the number of item quantity are 5.64 and 54.49 in average, respectively. Then, the true label of each instance is generated by classic column generation approach (e.g., which SPUs should be selected). To test the generality of our approach, the experiment is conducted on input orders and historical SPUs from different months, which can be seen in Table \ref{F1 result}.

\subsection{Experimental Setting}
In our MPN model, both the Process block and the Write block are  single-layer LSTM \cite{hochreiter1997long} with hidden size as 128. The selected features of SPU are set as Table.\ref{features table} and is mapped into a 128-dimension vector by two fully connected layers.   The batch size is set as 50. The model is implemented based on PyTorch \cite{paszke2019pytorch} and the optimizer is Adam \cite{KingmaB14} with the learning rate set as $10^{-4}$.

\subsection{F1 Score Result}

\begin{figure}[H]
  \centering
  \includegraphics[width=\linewidth]{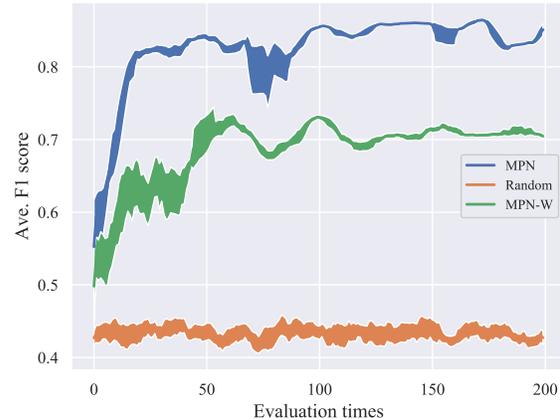}
  \caption{The F1 Score Curve.}
  \Description{The F1 score curve.}
  \label{curve}
\end{figure}

In this part, we present the expected reward curve during the training process as shown in Fig. \ref{curve} and the F1 score obtained from different months in Table. \ref{F1 result}. For ablation study, both the model with and without dynamic input mechanism are evaluated, respectively. 
First, in Fig. \ref{curve}, the MPN without dynamic input mechanism is denoted as MPN-W. In early stage, the MPN model arrives at a good performance more quickly compared to MPN-W. After 10 epochs training, the expected reward from MPN converges to a higher value than MPN-W. To verify the effectiveness of the proposed model, expected reward from random selecting is also presented. we can see that the value of F1 score from random selecting is around 0.45. 

Then, to test the generality, the comparison result on eight different input order sources are given in Table. \ref{F1 result}. Order source and historical records represent the month where input orders and historical SPUs are from, respectively. Both MPN and MPN-W models maintain outstanding performance during evaluation tests in all months, which means that the model trained based on orders from August is able to predict appropriate SPUs in all other months. Then, the F1 value obtained from MPN is higher than MPN-W, which means that MPN model dominates the MPN-W model and the dynamic input mechanism takes an significant role in simulating the problem status during the SPU selecting process of Write block.

\subsection{Packing Result}
In this part, we present the comparison result on packing success rate, the number of SPUs used, and the solving time.

First, the result on packing success rate is shown in Table. \ref{rate}.  The evaluation order source are from August to November 2020. For each month, the historical SPUs are from the previous month. For instance, input orders from August 2020 use historical SPUs from July 2020.

From the Table. \ref{rate}, in each month, the proposed data driven column generation approach gets better performance on packing success rate. As mentioned above, the fuzzy matching approach only selects the local best SPU at each iteration, which may lead to unsuccessful plan in some case. For instance, given an order $O_m=\left\{(A,2),(B,4),(C,1))\right\}$ and matched SPUs $SPU_1=\left\{(A,2),(B,3)\right\}$, $SPU_2=\left\{(A,1),(B,2)\right\}$, and $SPU_3=\left\{(C,1)\right\}$, the Fuzzy Match approach may select $SPU_1$ at the first stage. Then, in the second stage, all SPUs cannot be selected and the Fuzzy Match approach fails to handle this order. On the other hand, we can certainly obtain a feasible packing plan which includes two $SPU_2$ and one $SPU_3$ from the set cover formulation.
% \begin{table}[H]
% \caption{The Packing Rate}
% \label{rate}
% \begin{tabular}{ccccc}
% \toprule
% Order Source    & 8     & 9 & 10 & 11 \\ \midrule
% Fuzzy Match & 0.051 & 0.121 & 0.059  & 0.034  \\
% Column Generation       & 0.083 & 0.217 & 0.084  & 0.048  \\ \bottomrule
% \end{tabular}
% \end{table}

Then, in Table. \ref{packing result small}, we focus on the number of SPUs used and the solving time. In this experiment, at each month, we randomly select $2\times10^3$ orders successfully solved by classic column generation approach. The results are obtained with and without accelerating support by MPN, respectively.  In this table, the Ave.obj means that the average number of SPUs used and the total time denotes the total solving time for these tested orders. We can see that, in most of these months, the average  number of SPU used by MPN increases a little compared with that used by column generation approach. However, with the MPN accelerating, the solving time in all months decrease a lot. The result shows that our proposed MPN can select SPUs correctly and the predicting process consumes less time than the original column generation approach.

% \begin{table*}
%   \caption{Packing Quality and Solving Time }
%   \label{packing result small}
%  \begin{tabular}{ccccccccc}
%  \toprule
% Order Source             & \multicolumn{2}{c}{8} & \multicolumn{2}{c}{9} & \multicolumn{2}{c}{10} & \multicolumn{2}{c}{11} \\
% Metric            & Ave.obj       & Total.time      & Ave.obj       & Total.time      & Ave.obj       & Total.time      & Ave.obj        & Total.time      \\ \midrule
% Column Generation & 8.72                    & 71.35                    & 4.83                    & 70.25                    & 4.26                    & 60.21                    & 6.36                    & 54.84                    \\
% MPN                & 9.3                     & 44.86                    & 5.01                    & 56.96                    & 6.72                    & 57.7                     & 6.53                    & 47.19                       \\ \bottomrule
% \end{tabular}
% \end{table*}

\ \ To test the performance of our proposed approach in large scale orders (e.g., more types of items and the larger number of item quantity), based on $1.7\times10^5$ orders from August, by maintaining the total number of orders unchanged, we randomly combine $K$ input orders to generate a large scale dataset artificially.  With different values of $K$, the problem size is shown in Table. \ref{problem size}, in which the average number of item types, the average number of item quantity, and the average number of Matched SPUs are shown. In this part, we relax the constraint from customer's special requirements, which means the Matched SPU only need satisfy the quantity condition in Definition 1 and thereby the number of Matched SPUs naturally increases a lot. In this case, We only focus on the quantity relationship and test the ability of Fuzzy Match and our proposed approach for large scale problem.

\subsection{Large Scale Order}
\begin{figure}
  \centering
  \includegraphics[width=\linewidth]{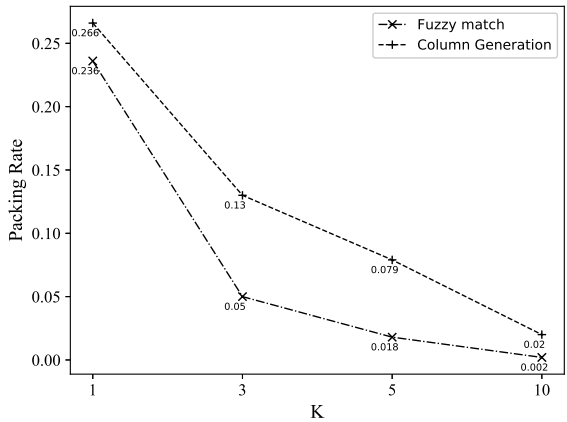}
  \caption{The Packing Success Rate for Large Scale Order.}
  \Description{The rate curve.}
  \label{large rate}
\end{figure}

\begin{table}[H]
    \centering
    \caption{Problem Size}
    \begin{tabular}{cccc}
    \toprule
    $K$& Type& Item &Matched Spus \\ \midrule
    1 &5.64&54.49&100.01 \\
    3 &18.24&175.09&319.46 \\
    5 &27.75&312.96&327.02 \\
    10 &58.17&596.25&369.69 \\ \bottomrule

    \end{tabular}
    
    \label{problem size}
\end{table}
In Fig. \ref{large rate}, we present the packing success rate with different values of $K$. First, $K=1$ represents that each individual order without combination is used for evaluation. The packing success rate increases a lot when the special demand constraint relaxes for both Fuzzy match and column generation. Then, with the value of $K$ becoming larger, the packing success rate decreases rapidly.  For the comparison result between Fuzzy Match and column generation approach, the packing success rate obtained from column generation approach always outperforms Fuzzy Match. The superiority of column generation approach becomes obvious as the value of $K$ increasing. When $k=10$, column generation approach even achieves ten times of packing success rate than Fuzzy Match. 
\begin{figure}
    \centering
    \includegraphics[width=\linewidth]{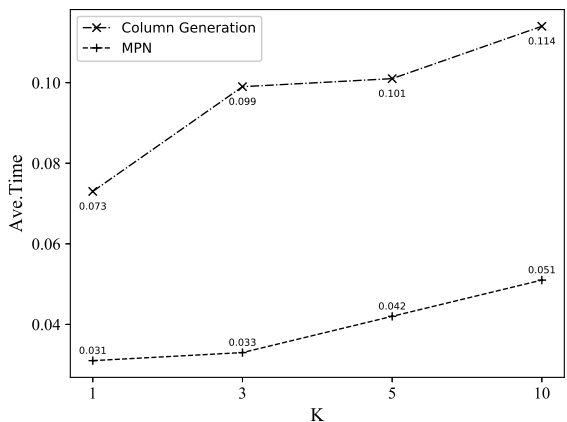}
    \caption{Comparison Result of Average Solving Time.}
    \label{time Comparison}
\end{figure}

\begin{figure}
    \centering
    \includegraphics[width=\linewidth]{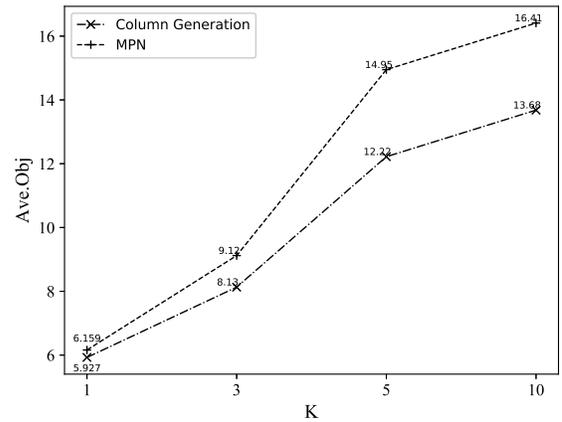}
    \caption{Comparison Result of Average Objective Value.}
    \label{obj Comparison}
\end{figure}
Then, the comparison result between column generation with and without MPN accelerating is shown in Fig. \ref{time Comparison} and Fig. \ref{obj Comparison}. For those orders successfully packed with column generation approach, we present the average solving time and average objective value, respectively. From these figures,  with the value of $K$ becoming larger, the solving time grows a lot due to the increase of problem size (e.g., the number of Matched SPUs and the type of items). Meanwhile, the large scale order needs more SPUs to be packed, thus, the value of objective also increases.  As for the solving time, the MPN always spends half or one third of time what has been spent by column generation approach. However, the number of SPUs used also becomes larger when the problem size becomes larger. When  $k=10$, the 
packing plan obtained from MPN uses two more SPUs than that by column generation approach in average, which meas that the Pointer Network loses ability of accurately selecting SPUs when the number of Matched SPUs is pretty large, e.g., the number of Matched SPUs is lager than 350.

\section{Conclusion}
In this paper, the historical data and the heuristic approach are integrated to reformulate the 3DBP problem as a set cover problem. A data driven column generation approach is proposed by incorporating the constraints handling strategy and the accelerating scheme. Rather than solving the pricing problem exactly, the constraints handling strategy takes utilization of historical data with evaluating the reduced cost of them, which is with the guaranteed feasibility under complex constraints, thus, the huge computational cost is avoided. Meanwhile, our proposed learning to price scheme accelerates the column generation by selecting SPUs from historical SPUs, and thus eliminates the iteration process. Through experiments on real-world datasets, we show our proposed method can improve the packing success rate and decrease the computation time simultaneously.

\begin{acks}
The corresponding authors are Dr. Zhenan He and Dr. Mingxuan Yuan.

\end{acks}

\bibliographystyle{ACM-Reference-Format}
\bibliography{reference}

% \appendix





\end{document}